\documentclass[]{article}

\newif\ifdraft
\usepackage[sort&compress]{natbib} 
\usepackage[lined,boxed,linesnumbered]{algorithm2e}
\usepackage{array}
\usepackage{amsmath}
 \usepackage{cite}
 \renewcommand{\cite}{\citep}
 
 \usepackage{endnotes}
 \usepackage[english]{babel}
 \usepackage[T1]{fontenc}
\usepackage[right]{lineno}
 \usepackage{multibib}
\usepackage{multicol} 
 \usepackage{multirow} 
 \usepackage{pifont}
 \usepackage{rotating} 
 \usepackage{soul} 
 \usepackage[table]{xcolor}
 \usepackage{tipa}
 \usepackage[utf8]{inputenc} 
 \usepackage[normalem]{ulem} 
\usepackage{url} 
\usepackage{tikz}
\newcommand*\correct[1]{\tikz[baseline=(char.base)]{
            \node[shape=circle,draw,ultra thick,inner sep=2pt] (char) {#1};}}
\newcommand*\wrong[1]{\tikz[baseline=(char.base)]{
            \node[shape=circle,draw,ultra thick,opacity=.5,inner sep=2pt] (char) {#1};}}

\newcommand{\cmark}{\ding{51}}
 \newcommand{\xmark}{\ding{55}}

\newcommand{\killpunct}[1]{} 
\newcolumntype{.}{D{.}{.}{-1}}
\newcolumntype{M}[1]{>{\centering\arraybackslash}m{#1}}
\newcolumntype{N}{@{}m{0pt}@{}}
\newcolumntype{C}[1]{>{\centering\let\newline\\\arraybackslash\hspace{0pt}}m{#1}}
\newcolumntype{Y}{>{\centering\arraybackslash}X}
\begin{document}

% \title{Is Dante Alighieri the author of the \\ ``Quaestio de Aqua et Terra''? \\
% A Study in \\ Computational Authorship Verification}

\title{The \textit{Questio de aqua et terra}: \\ A Computational
Authorship Verification Study}

% \author{Martina Leocata$^{1}$, Alejandro Moreo$^{1}$, Fabrizio Sebastiani$^{1}$, Marco Signori$^{2}$ \\ \mbox{} \\ $^{1}$ Istituto di Scienza e Tecnologie dell'Informazione \\
% Consiglio Nazionale delle Ricerche \\
% 56124 Pisa, Italy \\
% Email: \{firstname.lastname\}@isti.cnr.it \\ \mbox{} \\
% $^{2}$ Scuola IMT Alti Studi Lucca \\
% 55100 Lucca, IT \\
% Email: marco.signori@imtlucca.it \\
% \vspace{10ex} }

\author{Martina Leocata, Alejandro Moreo, Fabrizio Sebastiani \\ \mbox{} \\ $^{1}$ Istituto di Scienza e Tecnologie dell'Informazione \\
Consiglio Nazionale delle Ricerche \\
56124 Pisa, Italy \\
Email: \{\textit{firstname.lastname}\}@isti.cnr.it \\
\vspace{10ex} }

\date{}

\maketitle

% --------------------------------------------------------

\begin{abstract}
 \noindent 
%  \fabseb{Possible reviewers:
%  \begin{itemize}
%      \item Andreas van Cranenburgh
%      \item Paschalis Agapitos
%      \item Marco Passarotti
%      \item Justin Stover
%      \item Efstathios Stamatatos
%      \item Mike Kestemont
%      \item Jacques Savoy
%      \item Maciej Eder
%      \item Jan Rybicki
%  \end{itemize}
% }
 The \textit{Questio de aqua et terra} is a cosmological
 treatise traditionally attributed to Dante Alighieri. However, the
 authenticity of this text is controversial, due to discrepancies
 with Dante's established works and to the absence of contemporary
 references. This study investigates the authenticity of the
 \textit{Questio} via \emph{computational authorship verification}
 (AV), a class of techniques which combine supervised machine
 learning and stylometry. We build a family of AV systems and we
 assemble a corpus of 330 13th- and 14th-century Latin texts, which
 we use to comparatively evaluate the AV systems through
 leave-one-out cross-validation. Our best-performing system achieves
 high verification accuracy ($F_1=0.970$) despite the heterogeneity
 of the corpus in terms of textual genre. The key contribution to the
 accuracy of this system is shown to come from \textit{Distributional
 Random Oversampling} (DRO), a technique specially tailored to text
 classification which is here used for the first time in AV.
 
 The application of the AV system to the \textit{Questio} returns a
 highly confident prediction concerning its authenticity. These
 findings contribute to the debate on the authorship of the
 \textit{Questio}, and highlight DRO’s potential in the application
 of AV to cultural heritage.
\end{abstract}

% --------------------------------------------------------

% \tableofcontents

% --------------------------------------------------------

\section{Introduction}
\label{sec:introduction}

\noindent The \textit{Questio de aqua et terra} (literally: ``A
Question of the Water and of the Land'') is the text, allegedly
written by Dante Alighieri, of a lecture that Dante supposedly gave,
on January 20, 1320, in the Church of Saint Helen in Verona,
Italy. Written in Latin, this text was first published in 1520 by
Giovanni Benedetto Moncetti, a monk from Padova, Italy, who claimed to
have found the original manuscript, now unfortunately lost (as are the
original manuscripts of all of Dante's works, including the
\textit{Divine Comedy}). The \textit{Questio} is a treatise of a
cosmological nature, dealing with the interrelations among the four
fundamental elements (water, earth, air, and fire).

The authenticity of the \textit{Questio}, though, is still
debated. One of the reasons is that the cosmological views proposed in
it (a) seem to have originated in France and seem to have started
circulating in Italy only well after Dante's death, and (b)
inexplicably clash with the cosmological views espoused in the
\textit{Divine Comedy}. Another reason is that, while the other works
by Dante are mentioned and commented upon by many other 14th century
scholars, this is not the case for the \textit{Questio}, which seems
to have gone completely (and strangely) unnoticed throughout the 14th
and 15th centuries, with the sole exception of a mention found in a
version (whose authenticity is also disputed) of the \emph{Comment on
Dante's `Comedy'} by Pietro Alighieri (Dante's son). In sum, a number
of scholars 
% (see \citep{Casadei:2025yj, Fioravanti:2017ye}) 
\citep[see][]{Casadei:2025yj, Fioravanti:2017ye}
believe
that the lecture may never have been given, and that a yet unknown
forger (i) may have authored this text in order to espouse his own
cosmological views, and (ii) may have attributed it to Dante Alighieri
in order to grant credibility to these views. Unfortunately,
philologists who believe the \emph{Questio} not to be by Dante have
not given concrete suggestions as to who the real author might be.

In this work we investigate the authenticity of the \textit{Questio}
via the tools of \emph{computational authorship verification} (AV),
the task of building an automated system which checks whether a text
of controversial or unknown authorship was written by a certain
candidate author or not. In order to do so, (a) we first build a
corpus of 330 13th- and 14th-century texts written in Latin, (b) we
use this corpus to experimentally compare (via \emph{leave-one-out}
(LOO) testing, a maximally precise instance of $k$-fold
cross-validation) different AV systems (each consisting of the
combination of a supervised learning technique and a choice of feature
types), and (c) we apply to our disputed text the AV system that has
shown the best verification accuracy.

We show that our best-performing AV system exhibits, in a LOO test,
extremely high verification accuracy ($F_{1}=0.970$, corresponding to
329 correct inferences out of 330), despite the fact that the corpus
displays a high degree of heterogeneity in terms of literary genre. We
also show that our resulting authorship verifier issues a highly
confident prediction concerning the presumed Dantean authorship of the
\emph{Questio}.

Of particular interest is the fact that, as ablation experiments show,
the accuracy of the system is boosted by the use of
\emph{Distributional Random Oversampling} (DRO)~\citep{Moreo:2016hj}, 
a technique informed by the distributional
hypothesis \citep[see][]{Lenci:2023uz}
%\citep[see][]{Lenci:2023uz}
that generates, in a way
tailored to binary \emph{text} classification, artificial training
examples for the minority class. This fact is of special interest
since the present work is, to the best of our knowledge, the first
application of DRO (and of oversampling methods in general) to
authorship analysis tasks.

The paper is organized as follows. In order to make the paper
self-contained, in Section~\ref{sec:CAI} we briefly recap the basic
principles underlying computational authorship verification and the
supervised learning approach towards solving it. In Section
\ref{sec:AVfortheQuestio} we move to discussing how we have tackled AV
for the \emph{Questio}, detailing the construction of the corpus, the
supervised learning approach we adopt, and the various sets of
features we have tested. Section~\ref{sec:experiments} presents our
experimental results, which show how different combinations of a
supervised learning approach and various feature types behave, once
tested under the LOO protocol, on our corpus. Finally, in
Section~\ref{sec:Questio} we attempt to answer the question ``Is Dante
Alighieri the true author of the \emph{Questio}?'', and present
additional experiments by means of which we corroborate the
answer we return.
% Section~\ref{sec:relatedwork} discusses related work, while
Section~\ref{sec:conclusion} concludes by discussing how the present
work contributes to the debate on the authenticity of the
\textit{Questio}, and how DRO may prove a key tool in authorship
verification for cultural heritage.

In
order to allow our results to be verified by others, we make our
code and the \textsf{MedLatinQuestio} dataset publicly available.\footnote{\url{https://github.com/MartinaLeo/Questio_AV}} 
% \alex{\url{https://github.com/MartinaLeo/Questio_AV} è privato. Consiglierei modificare il readme (al momento vuoto) del repo per spiegare il contenuto.}

% -----------------------------------------------------------------

\section{Computational Authorship Identification (CAI)}
\label{sec:CAI}

\noindent How can we determine the author of a text of unknown or
disputed authorship? One way to do it is via an
\emph{intra-linguistic} analysis, based on the content or the style of
the text under investigation, i.e., based on ``endogenous'' (internal)
evidence. Another, complementary route is via an
\emph{extra-linguistic} analysis, based on mentions found in other
documents, on the historical context, or on the facts stated in the
text, i.e., based on ``exogenous'' (external) evidence. When
philologists investigate authorship via traditional means, they rely
on a \emph{qualitative} analysis of \emph{both} endogenous and
exogenous evidence. Computer scientists can instead contribute to
philological research by providing a \emph{quantitative} analysis,
based on endogenous evidence alone, through the means of
\emph{computational authorship identification} (CAI).
 
Two subtasks of CAI that are of interest to us are

\begin{itemize}
 
\item \emph{Authorship Attribution} (AA), the task of guessing, given
 a document $x$ and a set $\mathcal{A}$ of candidate authors to which
 the true author of $x$ is assumed to belong, who among the authors
 in $\mathcal{A}$ is the true author of $x$;
 
\item \emph{Authorship Verification} (AV), the task of guessing, given
 a document $x$ and a candidate author $A$, whether $A$ is the author
 of $x$ or not.
 
\end{itemize}

\noindent The basic assumption that underlies CAI is that authors tend
to leave their ``stylistic fingerprint'' on the texts they write, and
that this fingerprint may be detected computationally. The main focus
of the present work is AV (with Dante Alighieri as the candidate
author $A$ and the \textit{Questio} as the disputed text $x$), but we
will also recur to AA (see Section~\ref{sec:AA}) to corroborate the
results of our AV work.
 
% -----------------------------------------------------------------

\subsection{CAI as Text Classification}
\label{sec:CAIasclassification}

\noindent CAI tasks can be framed as \emph{text classification}
problems, which can be solved by applying supervised machine learning
techniques. In the supervised learning approach, we train a predictor
using a set $\mathcal{L} = \{(x_{i},y_{i})\}_{i=1}^{n}$ of examples of
known authorship (``training examples''), where $x_{i}$ denotes a text
and $y_{i}$ denotes the class to which it belongs. In our case, the
predictor can be an authorship verifier (in this case, if the
candidate author $A$ is Dante, the classes will be \texttt{Dante} and
\texttt{notDante}), or an authorship attributor (in this case, the
classes will each represent one of the candidate authors in
$\mathcal{A}$). In machine learning, AV corresponds to binary
classification, while AA corresponds to single-label multi-class
classification.

CAI combines supervised machine learning with \emph{stylometry}, the
quantitative analysis of textual style \citep{Kestemont:2014rw, 
Savoy:2020kt}; while
supervised machine learning provides the algorithms by means of which
we can train a predictor, stylometry provides the features that form
the vectorial representation used for conveying (a) the training
examples to the training algorithm and (b) the disputed text to the
trained predictor.
% In other words, machine learning provides the engine while
% stylometry provides the fuel.

% -----------------------------------------------------------------

\section{Authorship Verification for the \textit{Questio de aqua et
terra}}
\label{sec:AVfortheQuestio}

\noindent The main goal of our work is to train an authorship verifier
(i.e., a binary classifier) for classes \texttt{Dante} and
\texttt{notDante}, that we can use for determining whether the
\textit{Questio} is Dantean or not. Our work will consist of the
following four phases:

\begin{enumerate}
 
\item \label{item:corpus} Building the reference corpus, i.e., a set
 of documents of known authorship, some of which are known to be
 Dantean and the rest of which are known to be non-Dantean.
 
\item Choosing, via repeated experimentation, a ``maximally accurate''
 AV technique (i.e., an AV technique accurate to the best of our
 abilities), that we can later use for training, on the entire
 reference corpus, a verifier that we can then apply to the
 \textit{Questio}. This ``repeated experimentation'' consists of
 
 \begin{enumerate}
 
 \item \label{item:supl} Choosing a supervised machine learning
 algorithm;
 
 \item \label{item:rep} Choosing a stylometry-based vectorial
 representation of texts;
 
 \item \label{item:three} Training and testing an authorship verifier
 that uses the chosen representation and algorithm.
 
 \end{enumerate}
 
 In other words, we test (in Step \ref{item:three}) different
 combinations arising from choices made in Steps \ref{item:rep} and
 \ref{item:supl}, and we retain the best-performing one.
 
\item Applying this ``maximally accurate'' authorship verifier to the
 \textit{Questio}.
 
\item Analysing the reasons behind the verifier's determination
 concerning the \textit{Questio}.
 
\end{enumerate}
 
% -----------------------------------------------------------------

\subsection{Phase \ref{item:corpus}: Build the ``labelled''
(train+test) set}
\label{sec:buildthedataset}

\noindent We assemble a reference corpus consisting of a set of Latin
texts by (mostly Italian) authors from the 13th and 14th centuries. 

We start by including all the texts in the \textsf{MedLatinLit} corpus, a set, created and made available by
 \citet{Corbara:2022qv}, of 30 treatises and exegetical texts
 written in Latin by 13th-century Italian authors, for a total of 1,253,777 tokens (a ``token'' being either a punctuation symbol or a word); as the
examples of class \texttt{Dante}, the corpus includes Dante's only two
treatises, \textit{De Vulgari Eloquentia} and \textit{Monarchia}
(totalling 33,095 tokens altogether). 

We further add a corpus (which we call \textsf{MedLatinCosmo})\footnote{\label{foot:Signori}This set
was assembled on our behalf by Marco Signori, whom we sincerely
thank.} of 34 non-Dantean texts, written by other 13th or 14th century authors (e.g.,
Albertus Magnus, Robertus Grosseteste, Thomas de Aquino, etc.) and altogether
totalling 708,251 tokens; the texts are mostly treatises or
\textit{questiones} of a cosmological nature.\footnote{A
\textit{questio} (plural:\ \textit{questiones}) was an inquiry on a
specific issue, typically related to theological or philosophical
topics, carried out according to a method, often used in scholastic
thought, that involved a formal, structured process of analysis,
debate, and resolution.}
 
We further consider adding to our reference corpus

\begin{itemize}

\item the \textsf{MedLatinEpi} corpus, also created and made available
 by \citet{Corbara:2022qv}, a set of 264 epistles (i.e., letters) also
 written in Latin by 13th-century Italian authors (122,000 tokens), and including 12 Dantean epistles (6,859 tokens);

% \item the \textsf{MedLatinLit} corpus; created and made available by
%  \citet{Corbara:2022qv}, a set of treatises and exegetical texts
%  written in Latin by 13th-century Italian authors;

\item Dante's \textit{Eclogues}, two Latin poems in bucolic style (1,384 tokens).

\end{itemize}
\noindent While \textsf{MedLatinLit} and \textsf{MedLatinCosmo} consist of texts by and large pertaining to the same textual genre as the \textit{Questio}, \textsf{MedLatinEpi} and the \textit{Eclogues} do not, which means that their addition might be risky. However, preliminary experiments we have run show that, by adding them to \textsf{MedLatinLit} and
\textsf{MedLatinCosmo}, the accuracy of our
authorship verifiers on the latter two corpora increases, which means that, for the agoals of our task, \textsf{MedLatinEpi} and the \textit{Eclogues} contain ``more signal than noise''; we thus
merge all these four sources of text into what becomes our final
reference corpus, that we call \textsf{MedLatinQuestio}. The latter
eventually consists of 16 Dantean texts (for a total of 41,338 tokens)
and 314 non-Dantean texts (for a total of 2,044,344 tokens), and 
accounts for 38 different authors (including
Dante), with 15 authors represented by more than one text and 23
authors represented by one text only. In terms of textual genre,
\textsf{MedLatinQuestio} is heterogeneous, since it contains epistles,
texts of an exegetical nature, treatises, \textit{questiones} of a
cosmological nature, and even two poems. The full list of texts (with
their main characteristics) of which \textsf{MedLatinQuestio}
consists, is given in Appendix~\ref{sec:dataset}.

% -----------------------------------------------------------------

\subsection{Phase \ref{item:supl}: Define the supervised ML algorithm}
\label{sec:supervised}

% -----------------------------------------------------------------

\subsubsection{Logistic Regression}
\label{sec:logrec}

\noindent \noindent As the supervised learning algorithm we use
\emph{logistic regression} \citep[see e.g.,][\S6.4]{Aggarwal:2018dd}.
% (see e.g., \cite[\S6.4]{Aggarwal:2018dd}).
There are many reasons for this choice. First, logistic regression is
known to generate highly accurate classifiers when the items to
classify are texts \citep[see e.g.,][]{Aseervatham:2012on, Zhang:2003px, Genkin:2007ye}. Second, logistic regression is fast to train and
generates classifiers which are themselves fast; this is very useful
in our experiments, since leave-one-out experiments (see
Section~\ref{sec:LOO}) may become prohibitively demanding if the
training algorithm and the classifier are not efficient. Third, the
classifier trained by logistic regression is linear, which allows an
easy study of the relative importance that the different features have
had in verifying our disputed text \citep[see e.g.,][]{Setzu:2024qe}. Fourth, a classifier trained by logistic
regression returns, along with a binary prediction, a confidence score
(i.e., a measure of how confident the verifier is in the prediction it
has issued) in the form of a (``posterior'') probability; in other
words, the confidence score $\Pr(y|x)$ represents the probability that
the classifier attributes to the fact that the classified item $x$
belongs to class $y$.\footnote{Accordingly, $\Pr(y|x)=1$ (resp.,
$\Pr(y|x)=0$) indicates that the classifier is certain that $x$
belongs (resp., does not belong) to class $y$, while $\Pr(y|x)=0.50$
indicates that the classifier is completely uncertain as whether $x$
belongs to $y$ or not.} That confidence scores can be viewed as
probabilities is advantageous, since this makes these confidence
scores more easily interpretable. Last but not least, the posterior
probabilities returned by a classifier trained via logistic regression
tend to be well \emph{calibrated}, which (informally) means that they
are \emph{good-quality} posterior probabilities.
 
% -----------------------------------------------------------------

\subsubsection{Distributional Random Oversampling}
\label{sec:DRO}

\noindent Given the paucity of positive (i.e., Dantean) training texts,
% (only 121 segments \marleo{only 121 between segments and entire documents}), 
we resort to \textit{oversampling}. Oversampling
methods address binary classification scenarios in which the training
data are imbalanced (i.e., cases with much fewer examples of one class
than of the other) by generating additional (synthetic) positive
examples, already in vector form. The particular form of oversampling
we use is called \emph{distributional random oversampling} (DRO) 
\citep{Moreo:2016hj}. DRO is an oversampling method specifically
designed for classifying data, such as text, for which the
\emph{distributional hypothesis} holds 
\citep[see e.g.,][]{Lenci:2023uz}, a hypothesis stating that the meaning of a word
is somehow determined by the distribution of its occurrences in large
corpora of data. DRO extends the standard vectorial representation
with random, ``latent'' features based on distributional properties of
the existing features. In our application of DRO, each text 
% \marleo{rimuovere segment} 
is
associated to a discrete probabilistic function that can be queried as
many times as desired in order to oversample it (i.e., to
produce distributionally similar versions of it), where the
variability introduced in the newly generated examples reflects
semantic properties of the features that occur in the segment. See
\citep{Moreo:2016hj} for a detailed description of DRO.
 
% -----------------------------------------------------------------

\subsection{Phase \ref{item:rep}: Define the stylometry-based
vectorial representation of documents}
\label{sec:features}

\noindent We feed the training algorithm (and the final verifier, once
trained) with vectorial representations of our texts consisting of
several feature types, each encoding a different type of stylistic
information.

We choose the set of feature types to adopt via a greedy
\emph{iterative ablation} process, which consists of %
\begin{enumerate}

\item \label{item:start} choosing an initial pool
 $\mathcal{F}=\{F_{1}, ..., F_{m}\}$ of feature types;

\item \label{item:test} testing the verification accuracy $A(\mathcal{F})$
 obtained by using the entire set $\mathcal{F}$, for a chosen
 accuracy measure $A$;

\item for all $i\in\{1, ..., m\}$, testing the verification accuracy
 $A(\mathcal{F} \setminus \{F_{i}\})$ obtained by removing $F_{i}$ from
 pool $\mathcal{F}$;

\item defining
 $F^{*}\equiv\arg\displaystyle\max_{F_{i}\in\mathcal{F}}A(\mathcal{F}
 \setminus \{F_{i}\})$;

\item if $A(\mathcal{F} \setminus \{F^{*}\})\geq A(\mathcal{F})$,
 replacing $\mathcal{F}$ with $(\mathcal{F} \setminus \{F^{*}\})$ and
 repeating the cycle from Step~\ref{item:start}.

\end{enumerate}
\noindent The initial pool $\mathcal{F}$ of feature types we consider
is composed of the following types of features:

\renewcommand{\theenumi}{(\alph{enumi})}
\begin{enumerate}
 
\item \label{item:wls} Token lengths;
 
\item \label{item:} Function words, from a publicly available list of
 74 Latin function words used by \citet{Corbara:2022qv};
 
\item \label{item:sls} Sentence lengths, expressed as numbers of
 characters;
 
\item \label{item:POS} Part-of-speech (POS) $n$-grams (for
 $n\in\{1,2,3\}$);
 
\item \label{item:char} Character $n$-grams (for $n\in\{1,2,3\}$);

\item \label{item:syn} Syntactic dependency $n$-grams (for
 $n\in\{1,2,3\}$);

\item \label{item:morph} Morphosyntactyic verbal endings, from a
 publicly available list of 245 such endings used by
 \citet{Corbara:2022qv};

\item \label{item:mask} Features derived from character ``masking''
 (as developed by \citet{Stamatatos2018} -- see also \cite[\S
 3.3.2]{Corbara:2023kk}), for which we consider the DVMA and DVEX variants. 
 % \alex{Character "masking": dire quale tecnica, ce n'erano parecchie ma penso noi abbiamo solo usato DVEX o qualcuna simile.}
 
\end{enumerate}
\renewcommand{\theenumi}{\arabic{enumi}}

\noindent The reason why we use the values of $n$ indicated for feature types
\ref{item:POS}, \ref{item:char}, \ref{item:syn}, is that preliminary
experiments in which we have tested, for each of the three types of
features, values of $n$ higher than those, have not shown improvements
(or have shown a deterioration) in accuracy while, at the same time,
bringing about a substantial loss in efficiency, due to the high
number of features introduced.
Another type of features we have considered using is ``syllabic
quantities'' (as studied by \citet{Corbara:2023kk}), but we do not
attempt to use them (a) because the experiments run by
\citet{Corbara:2023kk} showed them to be ineffective when used on
medieval Latin, and because (b) their extraction and use involves a
considerable computational burden.
 
Note that these features types are fairly topic- and
genre-independent, which is important since we do not want our
features to act as ``confounding features'' \citep{Pearl:2009hi},
i.e., to allow the verifier to detect (i.e., be influenced by) topic
or genre instead of style alone. This is important since our dataset
is neither entirely topic-neutral (for instance, cosmology is fairly
prominent in the non-Dantean texts but not in the Dantean ones) nor
entirely genre-neutral (for instance, \textit{questiones} feature
promimently in the non-Dantean texts but not in the Dantean ones),
which entails the potential risk that the dataset biases the
authorship verifier towards a non-Dantean determination for the \textit{Questio}. It is
exactly by choosing topic- and genre-neutral features that one
counters this bias.
However, we observe that our features are only \textit{fairly} (and not
totally) topic- and genre-neutral; for instance, character 3-gram
``orb'' (as present in word ``orbis'', which is Latin for ``sphere'')
might feature more prominently in the negative (i.e., non-Dantean)
examples than in the positive (i.e., Dantean) ones because of the
prominence of cosmological texts in the former; this effect would be
even more pronounced if we used character $n$-grams for $n\geq 4$,
since the higher the value of $n$, the closer $n$-grams become to full
content-bearing words. Our choice of features thus attempts to strike
a good balance between complete topic- and genre-neutrality, and
informativeness.

Before generating our vectors, we carry out standard preprocessing; in particular, this includes lowercasing the text and then normalising it by 
exchanging (i) all occurrences of character \textit{v} with character
\textit{u} and (ii) all occurrences of character \textit{j} with
character \textit{i},\footnote{We do this in order to
standardise the different approaches that different editors might
follow. For example, in medieval written Latin, instead of the two
modern graphemes ``u-U'' and ``v-V'', there was only one grapheme,
represented as a lowercase ``u'' and a capital ``V''; some
contemporary editors follow this canon while others do not.} and by removing excerpts explicitly quoted from other texts. Prior to generating the vectors, we also
segment each of the 330 texts into
($\leq$400)-tokens-long non-overlapping
segments,\footnote{``($\leq$400)-tokens-long segments'' means that we
chop the text only at the end of full sentences, which means that a
segment is allowed to be longer than 400 tokens in order to allow a
sentence to be incorporated in the segment in full.} and also use the
resulting segments, along with the full texts, as training
examples. The rationale of this move is to increase the amount of
training examples, with the goal of countering the paucity of positive
training examples; in our case, segmentation results in 5,430 training
examples, of which 121 Dantean and 5,309 non-Dantean. Segmenting is a
standard pre-processing step in CAI, especially when dealing with
ancient texts, because these contexts are usually characterised by a
scarcity of training examples. The choice of 400 as the (lower bound
for the) length of our segments was the result of preliminary
experiments, in which, e.g., lengths such as 300 and 500 resulted in
inferior verification accuracy.
 
Each instance (segment or full text) is finally transformed into a
vector; for each feature the vector records its (TFIDF-weighted)
relative frequency of occurrence in the instance represented by the
vector.
 
% -----------------------------------------------------------------

\subsection{Phase 2c: Train and test the authorship verifier}
\label{sec:LOO}

\noindent In order to obtain a maximally accurate verifier, we test
different system configurations, resulting from either using or not
using (a) each of the feature sets listed in
Section~\ref{sec:features}, and (b) DRO. We test each configuration
via the \emph{leave-one-out} (LOO) protocol and choose the
configuration that displays the best verification accuracy. LOO
testing consists in repeating, independently for each text $x$ in the
labelled set $L$ (\textsf{MedLatinQuestio}, in our case) the cycle
consisting of
\begin{enumerate}

\item training the verifier,\footnote{This also includes computing the
 IDF component of TFIDF; this component is, as it should, recomputed
 anew in each iteration of the LOO protocol, so as not to involve the
 test document in its computation.} using the chosen system
 configuration,  (i) on all
 the texts in $L \setminus \{x\}$, and  (ii) on all the ($\leq$400)-tokens segments resulting from them (and on the artificial texts generated
 from (i)+(ii) by DRO, when DRO is used), 

\item applying the trained verifier on the (unsegmented) text $x$.

\end{enumerate}
\noindent As the measure for evaluating verification accuracy we use
the well-known $F_{1}$ measure, since it is the standard measure for
binary classification scenarios characterised (as our application is)
by imbalance. $F_{1}$ is a measure that mediates between
\emph{precision} (the ability of the system to avoid false positives)
and \emph{recall} (its ability to avoid false negatives), and is
defined as
\begin{align}
 \begin{split}
 \label{eq:F1}
 F_{1}=\frac{2\mathrm{TP}}{2\mathrm{TP}+\mathrm{FP}+\mathrm{FN}}
 \end{split}
\end{align}
\noindent
where TP, FP, FN, denote the numbers of true positives, false
positives, false negatives, resulting from the LOO test.
The LOO protocol has the advantage of (a) testing the accuracy of the
verification method on the largest possible set of texts (the entire
set $L$), which lends statistical robustness to the resulting accuracy
estimate, and (b) testing it on verifiers trained on approximately the
same number of training examples as the verifier that will be deployed
on the unlabelled text ($|L|-1$ for the former, $|L|$ for the latter),
which guarantees that the accuracy estimate is itself accurate. The
values of $F_{1}$ range between 0 (worst) and 1 (best).
 
Once we have run a LOO test for all the configurations we want to
test, we choose the configuration that has displayed the best accuracy
in the LOO test, use it to train a verifier on all the 330 texts in
\textsf{MedLatinQuestio}, and apply the trained verifier to the
\textit{Questio}. In case there is more than one best-performing
configuration in the LOO test, we choose the one that displays the
best value of \emph{soft $F_{1}$} (here indicated as $F_{1}^{s}$), a
measure similar to $F_{1}$ that we here define and whose goal is to
assess the quality not of the binary predictions but of the posterior
probabilities.\footnote{$F_{1}^{s}$ may be defined by considering the
values TP, FP, FN, TN of a binary contingency table not as counts but
as probability masses. E.g., while in the standard view a positive
example that has been correctly classified contributes 1 to the TP
cell and 0 to the FN cell, in this view a positive example such that
the posterior probability of the positive class is 0.80 contributes
0.80 to the TP cell and 0.20 to the FN cell; similarly, a negative
example such that the posterior probability of the positive class is
0.80 contributes 0.80 to the FP cell and 0.20 to the TN cell.
$F_{1}^{s}$ is computed via Equation~\ref{eq:F1} but using the contingency
table populated by the posterior probabilities instead of the one
populated by the binary predictions. $F_{1}^{s}$ is thus a measure of
the average degree of confidence that the system has in the correct
prediction (it is maximum when the system always makes the correct
prediction with confidence 1.00, i.e., when it is always the case that
$\Pr(y|x)=1$ and the true class of $x$ is $y$); unlike other such
measures, such as the Brier score, $F_{1}^{s}$ rewards systems that
balance their ability to avoid ``false negative
probability mass'' and their ability to avoid ``false positive
probability mass''.}

Note that, for each of the $|L|=330$ runs, we independently optimize
the $C$ hyperparameter of logistic regression, that defines the
tradeoff between training error and model complexity.

% -----------------------------------------------------------------

\section{Experimental results}
\label{sec:experiments}

\noindent A key part of our experiments is the greedy iterative ablation process  described in Section~\ref{sec:features}. The ideal way to carry it out would involve using, in  Step \ref{item:test}, a LOO test. Unfortunately, an iterative ablation process that uses LOO tests would be computationally too expensive (a single LOO test requires about one and a half days on a desktop computer equipped with a Intel Xeon Gold 6338 CPU @ 2.00GHz and 1TB of RAM, running Ubuntu 22). 
% \alex{Manca una descrizione della macchina in cui Martina ha girato gli sperimenti. Io spesso indico la mia cosi: "Times were clocked on a desktop computer equipped with a 12th Gen Intel(R) i9-12900K processor and 64GB of RAM, running Ubuntu 22.04.5"; aggiornare con quella giusta.} 
As a result, we carry out these tests in a more empirical way, i.e., we first identify the 10 texts that, using the initial pool $\mathcal{F}$ of feature types, are the hardest to classify correctly, and then use the classification accuracy obtained on these 10 texts as the guiding criterion for Step \ref{item:test}. 

In our experiments, the maximally accurate configuration 
turns out to be the one that uses
\begin{enumerate}

\item The set of feature types (identified via the iterative ablation
 process) consisting of (i)
 token lengths, for a total of 18 features;\footnote{The numbers of
 features mentioned here are those used in the final authorship
 verifier, trained on the entire \textsf{MedLatinQuestio} dataset. Of
 course, in each of the runs of which a leave-one-out test (see
 Section~\ref{sec:LOO}) consists, the numbers may be a bit smaller,
 and vary across runs; for instance, in the entire
 \textsf{MedLatinQuestio} dataset there are tokens of 18 different
 lengths, but in a specific run the number of different token lengths
 might be smaller since some token lengths might not be present in the
 training set of that run.} (ii) function words, for a total of 74
 features; (iii) sentence lengths, for a total of 998 features; (iv) POS
 $n$-grams (for $n\in\{1,2,3\}$), for a total of 3,346 features; (v)
 character $n$-grams (for $n\in\{1,2,3\}$), for a total of 8,371
 features. This generates a grand total of 12,807 features.

\item DRO. In our application of DRO we ask the system to generate a number of
synthetic positive examples such that the final ratio between positive
and negative examples is 20/80, a ratio usually considered optimal in
the literature on oversampling \citep[see][]{Moreo:2016hj}).\footnote{Note that, in our case study, DRO oversamples \textit{both} segments and full texts.} 
This
results in an extended, vectorized version of the
\textsf{MedLatinQuestio} dataset consisting of 6,636 examples (of
which 1,327 in class \texttt{Dante} and 5,309 in class
\texttt{notDante}), and described by 34,577 features (12,807
``natural'' features and 21,770 ``latent'' features generated by DRO).

\end{enumerate}
\noindent In the LOO experiments on our 330 texts this configuration
achieves a remarkable value of $F_1=0.970$.\footnote{The value of
vanilla accuracy, which is defined as the fraction of all
classification decisions that are correct, is instead
$\operatorname{VA}=329/330=0.997$.} 
% As summarised in
% Table~\ref{fig:resultsAV}, 
This results from the fact that the system
 
% \begin{table}[t]
%  \begin{center}
%  \begin{tabular}{|l|c|c|c|}
%  \cline{3-4}
%  \multicolumn{2}{c|}{} & \multicolumn{2}{c|}{True class} \\
%  \cline{3-4}
%  \multicolumn{2}{c|}{} & \texttt{Dante} & \texttt{notDante} \\
%  \hline
%  \multirow{2}{*}{Predicted class} & \texttt{Dante} & \cellcolor[HTML]{C6EFCE} 16 & \cellcolor[HTML]{FFC7CE} \phantom{31}1 \\
%  & \texttt{notDante} & \cellcolor[HTML]{FFC7CE} \phantom{1}0 & \cellcolor[HTML]{C6EFCE} 313 \\
%  \hline
%  \end{tabular}
%  \label{fig:resultsAV}
%  \caption{Contingency table illustrating the results obtained by
%  our authorship verifier, with 0 true negatives and just 1 true
%  positive.}
%  \end{center}
% \end{table}

\begin{itemize}
 
\item correctly evaluates as Dantean all the 16 Dantean texts
 (\textit{De Vulgari Eloquentia} + \textit{Monarchia} + 12 epistles +
 2 eclogues);
 
\item correctly evaluates as non-Dantean 313 out of the 314
 non-Dantean texts (it incorrectly evaluates Boccaccio's Epistle 23
 as Dantean).
 
\end{itemize}
\noindent The quality of the system is extremely good not only in
terms of the accuracy of its binary predictions, but also in terms of
the quality of its confidence scores, as witnessed by a $F_{1}^{s}$
score of 0.900. This high value of $F_{1}^{s}$ derives from the fact
that 321 out of 330 texts are correctly verified with confidence $>
.900$. The 10 texts hardest to verify are listed in
Table~\ref{fig:hardest}, which reports the probability that the system
attributes to the true class of the text. As shown in the table,
Giovanni Boccaccio is the author that the system considers
stylistically closer to Dante, since the system (a) erroneously
attributes to Dante one of his texts, and (b) classifies other 7 texts
by him as non-Dantean with some uncertainty.

\begin{table}[t]
 \begin{center}
 \begin{tabular}{|l|l||c|c|}
 \hline
 \multicolumn{1}{|c|}{\multirow{2}{*}{\textbf{Author}}} & 
 \multicolumn{1}{c||}{\multirow{2}{*}{\textbf{Text}}} & 
 \multicolumn{1}{c|}{\multirow{2}{*}{}} & 
 \multicolumn{1}{c|}{\multirow{1}{*}{\textbf{Confidence}}} \\
 & & & 
 \multicolumn{1}{c|}{\multirow{1}{*}{\textbf{in correct prediction}}} \\
 \hline
 Giovanni Boccaccio & Epistle 23 &  \xmark & 0.030 \\
 Giovanni Boccaccio & Epistle 16 &  \cmark & 0.575 \\
 Giovanni Boccaccio & Epistle 12 &  \cmark & 0.659 \\
 Giovanni Boccaccio & Epistle 6 &  \cmark & 0.683 \\
 Giovanni Boccaccio & Epistle 14 &  \cmark & 0.685 \\
 Giovanni Boccaccio & Epistle 17 &  \cmark & 0.734 \\
 Dante Alighieri & \textit{Monarchia} &  \cmark & 0.761 \\
 Giovanni Boccaccio & Epistle 1 &  \cmark & 0.837 \\
 Giovanni Boccaccio & Epistle 2 &  \cmark & 0.893 \\
 Guido Faba & Epistle 91 &  \cmark & 0.927 \\
 \hline
 \end{tabular}
 \label{fig:hardest}
 \caption{The texts that our authorship verifier finds the hardest
 to verify; the 3rd column indicates whether the system's inference
 proves correct (\cmark) or not (\xmark), while the 4th column
 indicates the system's confidence in the correct prediction (i.e.,
 the probability $\Pr(y|x)$ where $y$ is the true class of $x$).}
 \end{center}
\end{table}

% -----------------------------------------------------------------

\subsection{Which factors contributed most?}
\label{sec:whichfactors}

\noindent It may be interesting to measure how much each factor (e.g.,
use of a specific set of features, use of DRO, etc.) contributes to
the high level of accuracy of our authorship verifier. We do this via
factor \emph{ablation}, i.e., we run a LOO test using a system that
only differs from the maximally accurate configuration for the absence
of a specific factor. The results of these ablation experiments, one
for each set of features and one for DRO, are reported in
Table~\ref{fig:ablation}; in terms of $F_{1}$, they show that
\begin{table}[t]
 \centering 
 %\resizebox{\textwidth}{!} {
 \begin{tabular}{|l||c|c|c|c|}
 \hline
 \multicolumn{1}{|c||}{\multirow{2}{*}{System}} 
 & \multicolumn{1}{c|}{\multirow{2}{*}{FP}} 
 & \multicolumn{1}{c|}{\multirow{2}{*}{FN}} 
 & \multicolumn{1}{c|}{\multirow{2}{*}{$F_1$}}
 & \multicolumn{1}{c|}{\multirow{2}{*}{$F_{1}^{s}$}}
 \\
 & & & & \\
 \hline
 Maximally accurate configuration & 1 & ~0 & 0.970 & 0.900\\
 Without DRO & 0 & 12 & 0.400 & 0.480 \\
 Without token lengths & 5 & ~1 & 0.833 & 0.834\\ 
 Without function words & 5 & ~0 & 0.865 & 0.810 \\ 
 Without sentence lengths & 4 & ~1 & 0.857 & 0.834 \\ 
 Without POS $n$-grams & 6 & ~0 & 0.842 & 0.822 \\ 
 Without character $n$-grams & 1 & 11 & 0.455 & 0.431 \\ 
 \hline
 \end{tabular}
% }
 \label{fig:ablation}
 % \caption{The texts that proved hardest to verify for our
 % authorship verifier.}
 \caption{Results of our ablation experiments for AV.}
\end{table}
\begin{itemize}

\item The single most important factor that contributes to the high
 performance of the maximally accurate configuration is the use of
 DRO; without DRO, the system obtains a very disappointing $F_{1}$
 value of 0.400, due to the fact that no less than 12 out of 16
 Dantean texts are erroneously considered non-Dantean.

\item The other, almost equally important factor is character
 $n$-grams; without them, we again witness a dramatic drop in
 $F_{1}$, from 0.970 to 0.455, again caused by the fact that 11 out
 of 16 Dantean texts (plus one non-Dantean text) are misclassified.

\item The other four sets of features are way less critical to the
 performance of the verifier, since any system such that only one of
 them has been removed obtains an $F_{1}$ score higher than
 0.800. Still, each of them contributes non-negligibly.

\end{itemize}
 
% -----------------------------------------------------------------

\section{Is Dante the true author of the \textit{Questio}?}
\label{sec:Questio}

\noindent At this point, we are ready to ask our authorship verifier
whether Dante Alighieri is the true author of the \emph{Questio} or
not. In order to determine this, (a) we use the maximally accurate
configuration to train an authorship verifier on the full set of 330
texts of undisputed authorship, and (b) we apply the resulting
verifier to the \textit{Questio}. The result we obtain is that
\begin{enumerate}

\item according to our verifier, Dante is the true author of the
 \textit{Questio};

\item the verifier's degree of confidence in this fact (i.e., its
 posterior probability) is
 $\Pr(\mathtt{Dante}|\mathit{Questio})$=0.999999967; this is
 equivalent to saying that, according to the verifier, there are 33
 chances in a billion that the \textit{Questio} might be
 non-Dantean.\footnote{Since our maximally accurate configuration
 uses DRO, and since DRO extends the vector representing the
 unlabelled text with artificial features whose value it assigns
 nondeterministically, we use DRO to generate 10 ``random versions''
 of the \emph{Questio}, and classify each of them independently. In
 all of these 10 random runs the \emph{Questio} is classified as
 Dantean with a confidence higher than 0.999; the value of
 0.999999967 reported above is the median among these 10 values.}

\end{enumerate}
\noindent One question we may ask ourselves is if the almost absolute
certainty that the verifier has in the authenticity of the
\emph{Questio} may have been determined by the composition of the
corpus. We think that quite the opposite is true, since the
non-Dantean training examples are, on average, more similar (in terms
of topic and genre) to the \textit{Questio} than the positive
examples. Indeed, note that the set of non-Dantean training examples
contains several texts of a cosmological nature (which are thus
topically similar to the \textit{Questio}) and several
\textit{questiones} (which are thus similar to the \textit{Questio} in
terms of genre), while this is not true of the Dantean training
examples. In other words, if topic and genre had interfered with the
determination due to the imperfect topic- and genre-neutrality of the
features we have used (see the discussion in
Section~\ref{sec:features}), this should have pushed the system
towards stating that the \textit{Questio} is non-Dantean. Ultimately,
this means that the actual composition of the corpus even
\emph{reinforces} the credibility of the system's attribution of the
\textit{Questio} to Dante.

% -----------------------------------------------------------------

\subsection{Further investigations: Authorship attribution}
\label{sec:AA}

\noindent While the results we have just described seem to provide
strong evidence in favour of the Dantean paternity of the
\emph{Questio}, one may wonder whether we might obtain the same result
by subjecting the \textit{Questio} to an authorship verifier for one
of the other 38 candidate authors (e.g., whether an authorship
verifier for, say, Giovanni Boccaccio might claim that the
\emph{Questio} is by Boccaccio), thus generating a contradiction. A
thorough test of this hypothesis would be computationally expensive,
since we should not only train and apply to the \emph{Questio} 38
other authorship verifiers, but we should also run a LOO test for each
of these verifiers in order to estimate their reliability.

We circumvent the problem by subjecting the \textit{Questio} to a
system trained to perform \emph{authorship attribution} (see
Section~\ref{sec:CAI}), i.e., to predict who among a closed set of
candidate authors is, if at all, the true author of the
\textit{Questio}. To this end, we set up an AA system by (a)
considering as candidates the 15 authors for whom we have at least 2
texts (for the other 23 authors we have 1 text only),\footnote{A LOO
test of an authorship attribution system that involves an author $A$
for whom we have one text $x$ only would be vacuous; assuming, as it
should be, that the only text by this author is used as a training
example, we would have no test examples by this author, which means
that the ability of the system to attribute to $A$ the texts actually
written by $A$ could not be tested.} and (b) using the maximally
accurate configuration identified in our AV studies (logistic
regression + five feature sets, but without DRO, since DRO is a
technique that only applies to binary classification) to test via LOO
an AA system on the set of 307 texts written by one of these 15
authors. The LOO test returns a macroaveraged $F_1$ value (where
``macroaveraged $F_{1}$'' stands for the average $F_{1}$ across all
authors) of 0.628, resulting from correctly classifying 285 out of 307
texts and misclassifying the other 22.\footnote{The value of vanilla
accuracy is instead $\operatorname{VA}=285/307=0.928$.} This is a very
high macroaveraged $F_1$ value, since the task of picking, for each
item, the correct class among 15 different classes is a much more
difficult problem than choosing (as in AV) between 2 classes only.

We then train an AA system on the entire set of 307 texts and apply it
to the \emph{Questio}. The AA thus trained answers that, out of these
15 authors, the true author of the \textit{Questio} is Dante
Alighieri, with a probability of 0.900, which means that the system
attributes only a probability of 0.100 to the fact that the
\textit{Questio} is by one of the other 14 authors. The full
contingency table deriving from these experiments is reported as
Table~\ref{tab:confusionAA} in Appendix~\ref{app:AA}.
 
Still, we cannot exclude that one of the other 23 authors might have
been predicted to be the author of the \emph{Questio}, should we have
considered her/him as a candidate. 
% \alex{Mi spiegava Barbara Berti che in questi casi si scrive "should we have considered them".} 
% \fabseb{Sì, è uno dei modi che si usa in inglese per evitare il linguaggio sessista. Ma io lo trovo odioso, perché they/them per tutti noi è indissolubilmente legato al plurale, e questo uso rende il testo molto difficile da leggere. Io sono per lasciare her/him.} 
We thus train an AA system on the
entire set of 330 documents and the entire set of 38 authors, and
apply it to the \emph{Questio}. Also the AA thus trained answers that,
out of our 38 authors, the true author of the \textit{Questio} is
Dante, with a probability of 0.737, leaving a mere 0.263 estimated
probability that the \emph{Questio} might be by one of the other 38
authors (the 2nd top-ranked author, after Dante, is Antonio Pelacani
da Parma, with a probability of just 0.067). Table~\ref{tab:authors}
in Appendix~\ref{app:AA} lists the 38 authors in decreasing order of
probability that the AA system attributes to the fact that the author
is the person who actually wrote the \emph{Questio}.
 
Altogether, these results seem to exclude the possibility that an
authorship verifier for a different candidate author, when given the
\emph{Questio} to verify, might have answered affirmatively.
 
% -----------------------------------------------------------------

\subsection{Further investigations: Most similar texts}
\label{sec:mostsimilar}
 
\noindent In order to figure out the reasons why our AV and AA systems
so confidently attribute the \emph{Questio} to Dante, we rank the 330
texts in order of similarity to the \emph{Questio}, where the
similarity between two texts is defined as the cosine between the
vectors representing them, and where we use the same vectorial representations of
the texts as in our maximally accurate verifier of
Section~\ref{sec:experiments} (we do not use the artificial features inserted by DRO, though).
% \marleo{usiamo il setting senza DRO, perché altrimenti avremmo più copie dello stesso documento da confrontare}. 
The reason why we focus on similarity
to the \textit{Questio} is that, quite obviously, the texts recognised
as most similar to it are likely to play a major role in attributing
it to Dante.
 
Table~\ref{tab:mostsimilar} 
% (resp., Table~\ref{tab:leastsimilar})
lists the 10 texts that are found most similar 
% (resp., least similar)
to the \textit{Questio}. From them, we can see that
Dante's two treatises (which are also his two longest texts, i.e.,
those that give rise to the highest quantity of Dantean segments)
feature prominently, since his \textit{Monarchia} tops the chart of
the most similar texts, while his \textit{De Vulgari Eloquentia} is
ranked 7th out of 330.

% -----------------------------------------------------------------

% \subsection{The \textit{Questio} as a training example}
% \label{sec:questioastraining}
%
% \begin{itemize}
%
% \item When running LOO for authorship verification without DRO, 12
% Dantean texts (including the \textit{Monarchia}) are mistakenly
% classified as non-Dantean
%
% \item As an exercise, we have run the same LOO process but adding to
% the dataset the \textit{Questio} hypothetically labelled as
% Dantean
%
% \medskip
%
% \begin{scriptsize}
% \begin{tabular}{|l||c|c|c|c|}
% \hline
% \multicolumn{1}{|c||}{System} & FPs & FNs & $F_1$ & Soft $F_1$ \\
% \hline
% Maximally accurate configuration & 1 & \phantom{0}0 & 0.970 & 0.900 \\
% Without DRO & 0 & 12 & 0.400 & 0.480 \\
% Without DRO, with \textit{Questio} & 0 & \phantom{0}9 (out of 17) & 0.609 & 0.584 \\
% \hline
% \end{tabular}
% \end{scriptsize}
%
% \medskip
%
% \item As a result, 3 additional Dantean texts (including the
% \textit{Monarchia}, with confidence 0.999) are now correctly
% recognised as Dantean
%
% \item This seems to add further evidence towards the Dantean
% authorship of the \textit{Questio}
%
% \end{itemize}

% -----------------------------------------------------------------

\section{Discussion}
\label{sec:conclusion}
 
\noindent Our work provides strong evidence that Dante Alighieri may
be the true author of the \textit{Questio}. As the main sources of
evidence, we can list
\begin{enumerate}

\item The fact that our authorship verifier determines the
 \textit{Questio} to be Dantean, together with the fact that our
 authorship verifier's inferences are \emph{credible}. This
 credibility is witnessed by the fact that, when applied to a set of
 330 medieval Latin texts (many of which similar to the
 \textit{Questio} in terms of both topic and genre), it correctly
 determines 329 of them (which corresponds to $F_{1}=0.970$ and
 $\operatorname{VA}=0.996$). In particular, note that none of the
 texts similar to the \textit{Questio} in terms of either topic or
 genre, is misclassified by our verifier (the only misclassified
 document is an epistle).

\item The fact that our authorship verifier is extremely confident in
 the Dantean paternity of the \textit{Questio} (as indicated by
 $\Pr(\mathtt{Dante}|\mathit{Questio})$=0.999999967), together with
 the fact that our authorship verifier's degrees of confidence are
 \emph{reliable}, as shown by the $F_{1}^{s}$ value of $0.900$
 measured in the LOO test.

\item The fact that our authorship attribution system considers Dante
 by far the most likely author of the \textit{Questio} out of the 38
 authors we consider, with a probability of 0.737 (by contrast, the
 author considered the 2nd most likely obtains a probability of
 0.067, more than 10 times smaller).

\item The fact that, as discussed in Section~\ref{sec:Questio}, the
 very composition of the corpus could favour, if at all, a
 \textit{non}-Dantean determination, and certainly not a Dantean one.

\end{enumerate}
\noindent One question we might ask ourselves is whether this
determination could be overturned by other future experiments. In
principle, this might indeed happen. For instance, suppose that the
true author of the \textit{Questio} (let us call him, for convenience,
``Adso da Melk'') is not among our 38 authors, and that we happen to
obtain and add to our dataset a set of texts known to be by him. In
this case, it might well be that an authorship verifier tasked
with deciding whether the \emph{Questio} is by Adso da Melk or not
answers affirmatively, and that the authorship verifier tasked with
deciding whether the \emph{Questio} is by Dante or not answers
negatively (because of the fact that the non-Dantean training
documents have now undergone a substantial qualitative
change). However, we conjecture that, while possible, this hypothesis
is unlikely, since it would be difficult to explain why Dante has
stood out so prominently from the competition both in the AV and in
the AA experiments we have reported in Section~\ref{sec:Questio}. In
other words, if Adso from Melk is the true author, why does the AV
system of Section~\ref{sec:Questio} give such a strong Dantean
determination? why does the AA system of Section~\ref{sec:AA} indicate
that Dante is ten or one hundred times more likely to be the author of
the \textit{Questio} than all our other 38 authors?
 
It would be tempting to answer that this may happen because Adso da
Melk turns out to be a very good forger, and that he imitated Dante's
style so well as to deceive both the AV and the AA systems. However,
we think that, while possible, this hypothesis is unlikely too, since
the linguistic traits that we use as features are difficult, if not
impossible, to imitate. Indeed, the assumption that underlies
computational authorship analysis is that the linguistic
micro-phenomena that become the features of our vectorial
representations escape the conscious control of the author. In other
words, even if we assume that a skilled forger can successfully
imitate, say, Dante's sentence length frequencies (hard, but perhaps
not impossible), it seems virtually impossible that s/he can
imitate Dante's character $n$-gram frequencies. As a side note concerning
these latter, in an AV experiment that we run and that uses character
$n$-grams (for $n\in\{1,2,3\}$) as the \emph{only} set of features,
Dante is indicated as the true author of the \textit{Questio} with
probability 0.9988. 
% It is our impression that the ``forger theory''
% can hardly come to terms with this evidence. 
% \alex{Non so se sono d'accordo. Quello è un valore di posterior 
% probability molto alto, ma questo da solo non significa che il 
% sistema sia molto accurato; potrebbe dire 0.9988 per tutte le 
% opere.} \fabseb{Ho tolto l'ultima frase..}
 
From a technical standpoint, the most relevant fact that emerges from
our work is the \emph{strong contribution of DRO}, which brings $F_1$
from 0.400 to 0.970. To the best of our knowledge, DRO is being used
in AV (and in authorship analysis in general) for the very first time
(other aspects of our systems, such as the various feature types we
use, are instead not novel), and we believe that it holds promise for
computational authorship verification in cultural heritage, since the
latter is a context in which training texts are usually few, i.e., a
context in which we can benefit from the addition of \emph{plausible}
artificial training examples of the minority class.
 
Note that the improvement that DRO brought about was not granted. In
fact, DRO is based on the hypothesis (that is by now backed by a lot
of empirical evidence) that \emph{content-bearing words} obey the
distributional hypothesis. However, we have used DRO in an unorthodox
way, i.e., starting from vectorial representations of features
\emph{that are not content-bearing words}, but are function words,
%token lengths, 
POS $n$-grams, etc. 
% \marleo{il DRO è applicato solo ai feature set che restituiscono 
% matrici sparse, quindi su char n-grams, POS ngrams e function words, 
% non su sentence e token lengths} 
% \fabseb{C'è un motivo? E' impossibile usarlo su feature dense?}
In other words, the success that DRO
has had in extending vectorial representations of features that are
not content-bearing words somehow \emph{seems to suggest that also
these features obey the distributional hypothesis}, a fact which had
hardly been studied before and that might deserve further
investigation.

% -----------------------------------------------------------------

\section*{Acknowledgments}

\noindent We are indebted to Alberto Casadei for prompting us to apply
CAI techniques to the problem of the authorship of the
\textit{Questio}, and to Marco Signori for preparing the
\textsf{MedLatinCosmo} corpus, without which this work could not have
been carried out.

% -----------------------------------------------------------------

%\bibliographystyle{apalike} \bibliography{Fabrizio}

\begin{thebibliography}{}

\bibitem[Aggarwal, 2018]{Aggarwal:2018dd}
Aggarwal, C.~C. (2018).
\newblock {\em Machine learning for text}.
\newblock Springer International Publishing, Cham, CH.

\bibitem[Aseervatham et~al., 2012]{Aseervatham:2012on}
Aseervatham, S., Gaussier, {\'{E}}., Antoniadis, A., Burlet, M., and Denneulin,
  Y. (2012).
\newblock Logistic regression and text classification.
\newblock In Gaussier, {\'{E}}. and Yvon, F., editors, {\em Textual Information
  Access: Statistical Models}, pages 61--84. Wiley, Chichester, UK.

\bibitem[Casadei, 2025]{Casadei:2025yj}
Casadei, A. (2025).
\newblock Analysis of the 1509 edition and implications on the various
  reconstructions {[in Italian]}.
\newblock In Casadei, A. and Pontari, P., editors, {\em La \textit{Questio de
  aqua et terra}: Nuove indagini e prospettive}. Pisa University Press, Pisa,
  IT.
\newblock Forthcoming.

\bibitem[Corbara et~al., 2023]{Corbara:2023kk}
Corbara, S., Moreo, A., and Sebastiani, F. (2023).
\newblock Syllabic quantity patterns as rhythmic features for {L}atin
  authorship attribution.
\newblock {\em Journal of the Association for Information Science and
  Technology}, 74(1):128--141.

\bibitem[Corbara et~al., 2022]{Corbara:2022qv}
Corbara, S., Moreo, A., Sebastiani, F., and Tavoni, M. (2022).
\newblock {MedLatinEpi} and {MedLatinLit}: {Two} datasets for the computational
  authorship analysis of medieval {Latin} texts.
\newblock {\em ACM Journal of Computing and Cultural Heritage},
  15(3):57:1--57:15.

\bibitem[Fioravanti, 2017]{Fioravanti:2017ye}
Fioravanti, G. (2017).
\newblock {Alberto di Sassonia, Biagio Pelacani, and the «Questio de aqua et
  terra» [in Italian]}.
\newblock {\em Studi danteschi}, LXXXII:81--97.

\bibitem[Genkin et~al., 2007]{Genkin:2007ye}
Genkin, A., Lewis, D.~D., and Madigan, D. (2007).
\newblock Large-scale {Bayesian} logistic regression for text categorization.
\newblock {\em Technometrics}, 49(3):291--304.

\bibitem[Kestemont, 2014]{Kestemont:2014rw}
Kestemont, M. (2014).
\newblock Function words in authorship attribution: {F}rom black magic to
  theory?
\newblock In {\em Proceedings of the 3rd Workshop on Computational Linguistics
  for Literature (CLfL 2024)}, pages 59--66, Gothenburg, SE.

\bibitem[Lenci and Sahlgren, 2023]{Lenci:2023uz}
Lenci, A. and Sahlgren, M. (2023).
\newblock {\em Distributional semantics}.
\newblock Cambridge University Press, Cambridge, UK.

\bibitem[Moreo et~al., 2016]{Moreo:2016hj}
Moreo, A., Esuli, A., and Sebastiani, F. (2016).
\newblock Distributional random oversampling for imbalanced text
  classification.
\newblock In {\em Proceedings of the 39th ACM Conference on Research and
  Development in Information Retrieval (SIGIR 2016)}, pages 805--808, Pisa,
  {IT}.

\bibitem[Pearl, 2009]{Pearl:2009hi}
Pearl, J. (2009).
\newblock {Simpson's} paradox, confounding, and collapsibility.
\newblock In {\em {Causality: Models, Reasoning and Inference}}, chapter~6.
  Cambridge University Press, Cambridge, UK, 2nd edition.

\bibitem[Savoy, 2020]{Savoy:2020kt}
Savoy, J. (2020).
\newblock {\em Machine learning methods for stylometry: {A}uthorship
  attribution and author profiling}.
\newblock Springer, Cham, CH.

\bibitem[Setzu et~al., 2024]{Setzu:2024qe}
Setzu, M., Corbara, S., Monreale, A., Moreo, A., and Sebastiani, F. (2024).
\newblock Explainable authorship identification in cultural heritage
  applications.
\newblock {\em ACM Journal of Computing and Cultural Heritage}, 17(3):Article
  44.

\bibitem[Stamatatos, 2018]{Stamatatos2018}
Stamatatos, E. (2018).
\newblock Masking topic-related information to enhance authorship attribution.
\newblock {\em Journal of the Association for Information Science and
  Technology}, 69(3):461--473.

\bibitem[Zhang et~al., 2003]{Zhang:2003px}
Zhang, J., Jin, R., Yang, Y., and Hauptmann, A.~G. (2003).
\newblock Modified logistic regression: {An} approximation to {SVM} and its
  applications in large-scale text categorization.
\newblock In {\em Proceedings of the 12th International Conference on Machine
  Learning (ICML 2003)}, pages 888--895, Washington, US.

\end{thebibliography}

% -----------------------------------------------------------------

\newpage

\appendix

% -----------------------------------------------------------------

\section{Dataset}
\label{sec:dataset}

\noindent As discussed in Section~\ref{sec:buildthedataset}, our
\textsf{MedLatinQuestio} dataset consists of 4 sub-datasets, i.e., (i)
the \textsf{MedLatinCosmo} dataset that was assembled for this work
(see Footnote~\ref{foot:Signori}), (b) the \textsf{MedLatinEpi} and
\textsf{MedLatinLit} datasets made available by
\citet{Corbara:2022qv}, and (c) Dante Alighieri's two
\emph{Eclogues}. The texts that belong to each of these datasets are
listed in Tables~\ref{tab:MedLatinLit}, \ref{tab:MedLatinCosmo}, \ref{tab:MedLatinEpi},
\ref{tab:Eclogues}, along with their main
characteristics. 
% \fabseb{Qui ci vorrebbe una specifica delle edizioni
% critiche usate, sulla falsariga di quello che fece Silvia nel papero \citep{Corbara:2022qv}; ma forse la possiamo anche aggiungere nella versione finale del papero.}

\begin{table}[t]
 \caption{Composition of the \textsf{MedLatinLit} dataset; the 3rd
 column indicates the approximate historical period in which the
 text was written, while the 4th column indicates the number of tokens that the text consists of.  See \cite{Corbara:2022qv} for a
 specification of the critical editions used.
 % When assembled by
 % \citet{Corbara:2022qv}, the dataset included also Dante Alighieri's
 % \emph{De Vulgari Eloquentia} and \emph{Monarchia}, that we do not
 % list here since they are already part of \textsf{MedLatinCosmo}. 
 }
 \begin{scriptsize}
 \begin{center}
\bgroup \renewcommand{\arraystretch}{1.3}
 \resizebox{\textwidth}{!} {
 \begin{tabular}{|l|l|c|r|}
 \hline
 \multicolumn{1}{|c|}{\multirow{1}{*}{Author}} & \multicolumn{1}{c|}{\multirow{1}{*}{Text}} & Period & \multicolumn{1}{c|}{\multirow{1}{*}{\#Tokens}} \\
 \hline \hline
 Bene Florentinus & \textit{Candelabrum} & 1238 & 45,103 \\ \hline
  \multirow{3}{*}{Benvenuto Da Imola} & \textit{Comentum super Dantis Aldigherij Comoediam} & 1375-1380 & 117,361 \\
 & \textit{Glose Bucolicorum Virgilii} & 1380 & 4,184 \\
 & \textit{Expositio super Valerio Maximo} & 1380 & 3,199 \\
 \hline
  \multirow{4}{*}{Boncompagno Da Signa} & \textit{Liber de Obsidione Ancone} & 1198 & 9,248 \\
  & \textit{Palma} & 1204 & 5,924 \\
  & \textit{Rota Veneris} & $\leq$ 1215 & 2,108 \\
  & \textit{Ysagoge} & 1198-1200 & 4,732 \\
 \hline
  \multirow{2}{*}{Dante Alighieri} & \emph{De Vulgari Eloquentia} & 1304--1306 & 11,384 \\ 
  & \textit{Monarchia} & 1313--1319 & 19,162 \\ \hline
 Filippo Villani & \textit{Expositio seu Comentum super Comedia Dantis Allegherii} & 1391-1405 & 31,442 \\
 \hline
 \multirow{3}{*}{Giovanni Boccaccio} & \emph{De Vita et Moribus Domini Francisci Petracchi} & 1342 & 2,189 
 \\ 
 & \textit{De Mulieribus Claris} & 1361-1362 & 58,455 \\ 
 & \textit{De Genealogia Deorum Gentilium} & 1360-1375 & 220,142 \\ \hline
 \multirow{2}{*}{Giovanni del Virgilio} & \textit{Allegorie super Fabulas Ovidii Methamorphoseos} & 1320 & 23,095 \\ 
 & \emph{Ars Dictaminis} & 1320 & 2,530 \\ \hline
 Graziolo Bambaglioli & A Commentary on Dante's Inferno & 1324 & 41,104 \\ \hline
 Guido of Pisa & \textit{Expositiones et Glose. Declaratio super Comediam Dantis} & 1327-1328 & 97,305 \\ \hline
 Guido de Columnis & \textit{Historia Destructionis Troiae} & 1272-1287 & 93,694 \\ \hline
 Guido Faba & \textit{Dictamina Rhetorica} & 1226-1228 & 19,467 \\ \hline
 Jacobus de Voragine & \textit{Chronica Civitatis Ianuensis} & 1295-1298 & 62,589 \\ \hline
 Jean d'Eppe & \textit{Constitutiones Romandiolae} & 1283 & 4,483 \\ \hline
 Giovanni da Pian del Carpine & \textit{Historia Mongalorum} & 1247-1252 & 23,557 \\ \hline
 Julian of Speyer  & \textit{Vita Sancti Francisci} & 1232-1239 & 14,432 \\ \hline
 \multirow{2}{*}{Nicholas Trevet } & \emph{Expositio Herculis Furentis} & 1315-1316 & 40,014 \\ 
 & \textit{Expositio L.\ Annaei Senecae Agamemnonis} & 1315-1316 & 24,833 \\ \hline
 Pietro Alighieri & \textit{Comentum super Poema Comedie Dantis} & 1340-1364 & 135,320 \\ \hline
 Ramon Llull & \textit{Ars Amativa Boni} & 1290 & 98,181 \\ \hline
  Richard of San Germano  & \textit{Chronicon} & 1216-1243 & 35,761 \\ \hline
 Zono de’ Magnalis & \textit{Vita Virgilii} & 1340 & 2,297 \\ \hline\hline
 \multicolumn{3}{|r}{Total number of tokens $\rightarrow$}  & 1,253,777 \\ \hline
 \end{tabular}
 }
 \egroup
 \end{center}
 \end{scriptsize}
 \label{tab:MedLatinLit}
\end{table}%

 \begin{table}[h!]
 \caption{Composition of the \textsf{MedLatinCosmo} dataset; the
 meanings of the columns are as in
 Table~\ref{tab:MedLatinLit}.}
 \begin{footnotesize}
 \begin{center}
 \bgroup \renewcommand{\arraystretch}{1.3}
 \resizebox{\textwidth}{!} {
 \begin{tabular}{|l|l|c|r|}
 \hline
 \multicolumn{1}{|c|}{\multirow{2}{*}{Author}} & \multicolumn{1}{c|}{\multirow{2}{*}{Text (or collection thereof)}} & Period & \multicolumn{1}{c|}{\multirow{2}{*}{\#Tokens}} \\
 & & (approx.) & \\
 \hline \hline
 \multirow{4}{*}{Albertus Magnus} & \textit{De Caelo et Mundo} (Book I -- Treatise I) & 1251-1254 & 18,726 \\
 % \cline{2-5}
 & \textit{Meteora} (Book I -- Treatises I,II,III,IV) & 1254-1257 & 19,437 \\
 & \textit{Liber de Principiis Motus Processivi} & 1258-1263 & 16,827 \\
 & \textit{De Unitate Intellectus} & 
 %“Endfassung” 
 1263 & 19,077 \\
 \hline
 Antonio Pelacani of Parma & Comment on Avicenna's Canon & 1310-1323 & 165,810 \\
 \hline
 Bartholomeus de Bononia
 % Bartholomew of Bologna
 & \textit{De Luce} & 1270 & 22,422 \\
 \hline
 Cicchus Esculanus & \textit{Commentum in Iohannis de Sacrobosco} & $\approx$ 1300 & 29,983 \\
 \hline
  \multirow{4}{*}{Girardus Cremonensis}  & \textit{De Somno et Visione} & $\leq$ 1187 & 3,543 \\
 & \textit{Farabi} & $\leq$ 1187 & 16,665 \\
 & \textit{Fontes Quaestionum} & $\approx$ 1200 & 972 \\
 & \textit{Liber de Diffinitionibus} & $\leq$ 1187 & 7,761 \\
 \hline
  \multirow{2}{*}{Guido Vernani}  & \textit{Adbreviatio Expositionis} & $\approx$ 1300 & 4,588 \\
 & \textit{De Rebrobatione Monarchie} & 1327-1334 & 8,108 \\
 \hline
 Iohannes De Sacrobosco & \textit{Tractatus de Spera} & 1220-1230 & 9,086 \\
 \hline
 Iohannes Peckham & \textit{Tractatus de Spera} & $\approx$ 1250 & 2,272 \\
 \hline
 Robert Kilwardby & \textit{De Natura Relationis} & $\leq$ 1279 & 18,743 \\
 \hline
 Michael Scotus & \textit{Eximi atque} & $\approx$ 1300 & 32,978 \\
 \hline
  \multirow{2}{*}{Nicole Oresme}  & \textit{Questiones in Meteorologica Liber} I & 1340-1356 & 35,581 \\
 & \textit{Questiones in Meteorologica Liber} II & 1340-1356 & 15,803 \\
 \hline
 Ptolemaeus Lucensis & \textit{Determinatio Compendiosa} & $\leq$ 1327 & 15,360 \\
 \hline
 Robertus Anglicus & \textit{Glossa in Tractatu de Sphera} & 1271 & 20,257 \\
 \hline
 \multirow{3}{*}{Robertus Grosseteste}  & \textit{De Cometis} & 1222-1224 & 1,532 \\
 & \textit{De Motu Supercelestium} & $\approx$ 1225 & 3,191 \\
 & \textit{De Sphaera} & 1216-1220 & 6,234 \\
 \hline
  \multirow{9}{*}{Thomas De Aquino}  & \textit{De Caelo et Mundo} & 1272-1273 & 56,399 \\
  & \emph{Questiones Quodlibetales} I 4 1 & 1256-1259 & 1,038 \\
  & \emph{Questiones Quodlibetales} II 2 2 & 1256-1259 & 1,331 \\
  & \emph{Questiones Quodlibetales} III 1 2 & 1256-1259 & 411 \\
  & \emph{Questiones Quodlibetales} III 14 1 & 1256-1259 & 572 \\
  & \emph{Questiones Quodlibetales} III 4 1 & 1256-1259 & 719 \\
  & \emph{Questiones Quodlibetales} VI 11 & 1256-1259 & 782 \\
  & \textit{Sentencia Libri Politicorum} & 1269-1272 & 66,063 \\
  & \textit{Sentencia super Meteora} & 1269-1273 & 40,476 \\
 \hline
 William of Moerbeke & Ptolemy
\textit{Quadripartitum} & $\approx$ 1200 & 35,815 \\
\hline\hline
 \multicolumn{3}{|r}{Total number of tokens $\rightarrow$}  & 708,521 \\ 
 \hline\end{tabular}
 }
 \egroup
 \end{center}
 \end{footnotesize}
 \label{tab:MedLatinCosmo}
 \end{table}

\begin{table}[t]
 \caption{Composition of the \textsf{MedLatinEpi} dataset; the 3rd
 column indicates the approximate historical period in which the
 texts were written, while the 4th and 5th columns indicate the number of
 texts and the number of tokens that the collection consists of. Differently from
 \citet{Corbara:2022qv}, we do not include the 30 epistles from the
 collection of Petrus de Boateriis, since they are by miscellaneous
 authors and, especially in the AA experiments, they would thus bring about
 more problems than solutions. See \cite{Corbara:2022qv} for a
 specification of the critical editions used.}
 \begin{footnotesize}
 \begin{center}
 \bgroup \renewcommand{\arraystretch}{1.3}
 \resizebox{\textwidth}{!} {
 \begin{tabular}{|l|l|c|r|r|}
 \hline
 \multicolumn{1}{|c|}{\multirow{2}{*}{Author}} & \multicolumn{1}{c|}{\multirow{2}{*}{Text (or collection thereof)}} & Period & \multicolumn{1}{c|}{\multirow{2}{*}{\#Texts}} & \multicolumn{1}{c|}{\multirow{2}{*}{\#Tokens}} \\
 & & (approx.) & & \\
 \hline \hline
 \multirow{2}{*}{Clare of Assisi} & \textit{Epistola ad Ermentrudem} & 1240-1253 & 1 & 298 \\
 % \cline{2-5}
 & \textit{Epistolae ad Sanctam Agnetem de Praga} I, II, III & 1234-1253 & 3 & 2,144 \\
 \hline
 Dante Alighieri & Epistles & 1304-1315 & 12 & 6,859 \\
 \hline
 Giovanni Boccaccio & Epistles and Letters & 1340-1375 & 24 & 28,894     \\ 
 \hline
 Guido Faba & Epistles & 1239-1241 & 78 & 8,122 \\
 \hline
 Pietro della Vigna & The Collected Epistles of Pietro della Vigna & 1220-1249 & 146 & 75,683 \\ 
\hline\hline
 \multicolumn{4}{|r}{Total number of tokens $\rightarrow$}  & 122,000 \\ 
 \hline
 \end{tabular}
 }
 \egroup
 \end{center}
 \end{footnotesize}
 \label{tab:MedLatinEpi}
 \end{table}

% -----------------------

 \begin{table}[t]
 \caption{Dante Alighieri's two \emph{Eclogues}; the meanings of the
 columns are as in Table~\ref{tab:MedLatinLit}. }
 \begin{footnotesize}
 \begin{center}
 \bgroup \renewcommand{\arraystretch}{1.3}
 \resizebox{\textwidth}{!} {
 \begin{tabular}{|l|l|c|r|}
 \hline
 \multicolumn{1}{|c|}{\multirow{2}{*}{Author}} & \multicolumn{1}{c|}{\multirow{2}{*}{Text (or collection thereof)}} & Period  & \multicolumn{1}{c|}{\multirow{2}{*}{\#Tokens}} \\
 & & (approx.) & \\
 \hline \hline
 Dante Alighieri & \textit{Vidimus in Nigris Albo Patiente Lituris} & 1319-1320 & 598 \\
 Dante Alighieri & \textit{Velleribus Colchis Prepes Detectus Eous} & 1319-1320 & 786 \\
\hline\hline
 \multicolumn{3}{|r}{Total number of tokens $\rightarrow$}  & 1,384 \\ 
 \hline
 \end{tabular}
 }
 \egroup
 \end{center}
 \end{footnotesize}
 \label{tab:Eclogues}
\end{table}

\section{Authorship attribution results}
\label{app:AA}

\noindent Table~\ref{tab:confusionAA} reports, in the form of a
contingency table, the results of our authorship attribution
experiments on the 15 authors for which the \textsf{MedLatinQuestio}
corpus contains at least 2 texts.
 
The results seem to suggest that some authors are ``close'' to each
other in terms of style, as witnessed by the fact that the verifier
has a tendency to attribute to one author the texts of the other,
and/or vice versa. For instance, some closeness in style seems to
exist between Boncompagno da Signa and Dante Alighieri, as indicated
by the fact that the verifier attributes 2 texts by the former to the
latter, and 1 text of the latter to the former. Another example is a
detected closeness between Robertus Grosseteste and Albertus Magnus,
as shown by the fact that all 3 texts by the former are mistakenly
attributed to the latter; similar observations seem to indicate some
degree of similarity between Pietro della Vigna and Benvenuto da
Imola. Other authors such as Girardus Cremonensis, Nicholas Trevet,
Nicole Oresme, and Thomas De Aquino, seem instead to have a truly
unique style, since none of their texts is misattributed to others and
no text by others is misattributed to them.
 
Table~\ref{tab:authors} lists the 38 authors represented in the
\textsf{MedLatinQuestio} corpus in decreasing order of probability
that the AA system attributes to the fact that the author is the one
who actually wrote the \emph{Questio}.
 
\begin{table}[htb]
 \begin{center}
 \bgroup \renewcommand{\arraystretch}{1.38}
 \resizebox{.90\textwidth}{!} {
 \begin{tabular}{|c|l|ccccccccccccccc|c|}
 \cline{3-17}
 \multicolumn{2}{c|}{} & \multicolumn{15}{c|}{Predicted Author} \\
 \cline{3-17}
 \multicolumn{2}{c|}{} & \rotatebox{90}{Albertus Magnus} & \rotatebox{90}{Benvenuto da Imola} & \rotatebox{90}{Boncompagno da Signa~} & \rotatebox{90}{Clare of Assisi} & \rotatebox{90}{Dante Alighieri} & \rotatebox{90}{Giovanni Boccaccio} & \rotatebox{90}{Giovanni Del Virgilio} & \rotatebox{90}{Girardus Cremonensis} & \rotatebox{90}{Guido Vernani} & \rotatebox{90}{Guido Faba} & \rotatebox{90}{Nicholas Trevet} & \rotatebox{90}{Nicole Oresme} & \rotatebox{90}{Pietro della Vigna} & \rotatebox{90}{Robertus Grosseteste} & \rotatebox{90}{Thomas De Aquino} \\
 \hline
 \multirow{15}{*}{\rotatebox{90}{True Author}} 
 & Albertus Magnus & \correct{4} & 0 & 0 & 0 & 0 & 0 & 0 & 0 & 0 & 0 & 0 & 0 & 0 & 0 & 0 \\
 & Benvenuto da Imola & 0 & \correct{3} & 0 & 0 & 0 & 0 & 0 & 0 & 0 & 0 & 0 & 0 & 0 & 0 & 0 \\
 & Boncompagno da Signa~ & 0 & 0 & \correct{2} & 0 & \wrong{2} & 0 & 0 & 0 & 0 & 0 & 0 & 0 & 0 & 0 & 0 \\
 & Clare of Assisi & 0 & \wrong{1} & 0 & 0 & 0 & \wrong{1} & 0 & 0 & 0 & 0 & 0 & 0 & \wrong{2} & 0 & 0 \\
 & Dante Alighieri & 0 & 0 & \wrong{1} & 0 & \correct{13} & \wrong{1} & 0 & 0 & 0 & \wrong{1} & 0 & 0 & 0 & 0 & 0 \\
 & Giovanni Boccaccio & 0 & 0 & 0 & 0 & 0 & \correct{26} & 0 & 0 & 0 & 0 & 0 & 0 & 0 & 0 & 0 \\
 & Giovanni Del Virgilio & 0 & \wrong{1} & \wrong{1} & 0 & 0 & 0 & 0 & 0 & 0 & 0 & 0 & 0 & 0 & 0 & 0 \\
 & Girardus Cremonensis & 0 & 0 & 0 & 0 & 0 & 0 & 0 & \correct{4} & 0 & 0 & 0 & 0 & 0 & 0 & 0 \\
 & Guido Vernani & \wrong{1} & 0 & 0 & 0 & \wrong{1} & 0 & 0 & 0 & 0 & 0 & 0 & 0 & 0 & 0 & 0 \\
 & Guido Faba & 0 & 0 & \wrong{1} & 0 & 0 & 0 & 0 & 0 & 0 & \correct{76} & 0 & 0 & \wrong{2} & 0 & 0 \\
 & Nicholas Trevet & 0 & 0 & 0 & 0 & 0 & 0 & 0 & 0 & 0 & 0 & \correct{2} & 0 & 0 & 0 & 0 \\
 & Nicole Oresme & 0 & 0 & 0 & 0 & 0 & 0 & 0 & 0 & 0 & 0 & 0 & \correct{2} & 0 & 0 & 0 \\
 & Pietro della Vigna & 0 & \wrong{3} & 0 & 0 & 0 & 0 & 0 & 0 & 0 & 0 & 0 & 0 & \correct{143} & 0 & 0 \\
 & Robertus Grosseteste & \wrong{3} & 0 & 0 & 0 & 0 & 0 & 0 & 0 & 0 & 0 & 0 & 0 & 0 & 0 & 0 \\
 & Thomas De Aquino & 0 & 0 & 0 & 0 & 0 & 0 & 0 & 0 & 0 & 0 & 0 & 0 & 0 & 0 & \correct{9} \\
 \hline
 \end{tabular}
 }
 \egroup
 \caption{Contingency table deriving from our authorship attibution
 experiments on the 15 authors for which the
 \textsf{MedLatinQuestio} corpus contains at least 2 texts; correct
 attributions are displayed in black circles while misattributions are
 displayed in grey circles.}
 \label{tab:confusionAA}
 \end{center}
\end{table}

\begin{table}[htbp]
 \bgroup \renewcommand{\arraystretch}{1.3}
 \begin{center}
 \begin{tabular}{cc}
 \resizebox{.48\textwidth}{!} {
 \begin{tabular}{|rlc|}
 \hline
 \# & \multicolumn{1}{c}{Author} & Probability \\
 \hline
% \phantom{0} & \phantom{Giovanni da Pian Del Carpine} & \phantom{Probability} \\
 1 & Dante Alighieri & 0.7371 \\
 2 & Antonio Pelacani da Parma\hspace{2.8ex} & 0.0672 \\
 3 & Pietro Alighieri & 0.0491 \\
 4 & Robertus Grosseteste & 0.0190 \\
 5 & Nicole Oresme & 0.0183 \\
 6 & Cicchus Esculanus & 0.0144 \\
 7 & Michael Scotus & 0.0139 \\
 8 & Robert Kilwardby & 0.0137 \\
 9 & Bartholomeus de Bononia & 0.0119 \\
 10 & Robertus Anglicus & 0.0110 \\
 11 & Guido Vernani & 0.0067 \\
 12 & Graziolo Bambaglioli & 0.0036 \\
 13 & Giovanni Del Virgilio & 0.0035 \\
 14 & Girardus Cremonensis & 0.0034 \\
 15 & Guido of Pisa & 0.0030 \\
 16 & Guido De Columnis & 0.0030 \\
 17 & Filippo Villani & 0.0028 \\
 18 & Iohannes Peckham & 0.0025 \\
 19 & Ramon Llull & 0.0021 \\
 \hline
 \end{tabular}
 }
 \resizebox{.48\textwidth}{!} {
 \begin{tabular}{|rlc|}
 \hline
 \# & \multicolumn{1}{c}{Author} & Probability \\
 \hline
% \phantom{0} & \phantom{Giovanni da Pian Del Carpine} & \phantom{Probability} \\
 20 & Albertus Magnus & 0.0021 \\
 21 & Thomas De Aquino & 0.0019 \\
 22 & Ptolemaeus Lucensis & 0.0017 \\
 23 & Jacobus de Voragine & 0.0014 \\
 24 & Iohannes De Sacrobosco & 0.0013 \\
 25 & Giovanni da Pian Del Carpine & 0.0009 \\
 26 & Boncompagno Da Signa & 0.0008 \\
 27 & Benvenuto Da Imola & 0.0007 \\
 28 & Bene Florentinus & 0.0006 \\
 29 & Julian of Speyer & 0.0004 \\
 30 & Nicholas Trevet & 0.0003 \\
 31 & Zono de' Magnalis & 0.0003 \\
 32 & Richard of San Germano & 0.0003 \\
 33 & William of Moerbeke & 0.0001 \\
 34 & Clare of Assisi & 0.0001 \\
 35 & Jean d'Eppe & 0.0000 \\
 36 & Giovanni Boccaccio & 0.0000 \\
 37 & Guido Faba & 0.0000 \\
 38 & Pietro della Vigna & 0.0000 \\
 \hline
 \end{tabular}
 }
 \end{tabular}
 \end{center}
 \caption{Our 38 authors, listed in decreasing order of probability
 that the AA system attributes to the fact that the author is the one
 who actually wrote the \emph{Questio}.}
 \label{tab:authors}
 \egroup
\end{table}

\begin{table}[htp]
 \begin{center}
 \bgroup \renewcommand{\arraystretch}{1.3}
 \resizebox{\textwidth}{!} {
 \begin{tabular}{|rllc|}
 \hline
 \multicolumn{2}{|c}{Author} & \multicolumn{1}{c}{Text} & \multicolumn{1}{c|}{Similarity} \\
 \hline
 1 & Dante Alighieri & Monarchia & 0.9344 \\
 2 & Nicole Oresme & Questiones in Meteorologica Liber II & 0.9337 \\
 3 & Michael Scotus & Eximi Atque & 0.9290 \\
 4 & Bartholomeus de Bononia & De Luce & 0.9268 \\
 5 & Guido Vernani & De Rebrobatione Monarchie & 0.9263 \\
 6 & Nicole Oresme & Questiones in Meteorologica Liber I & 0.9261 \\
 7 & Dante Alighieri & De Vulgari Eloquentia & 0.9256 \\
 8 & Albertus Magnus & De Meteora & 0.9236 \\
 9 & Robertus Anglicus & Glossa in Tractatu de Sphera & 0.9235 \\
 10 & Robert Kilwardby & De Natura Relationis & 0.9210 \\
 \hline
 \end{tabular}
 }
 \egroup
 \end{center}
 \caption{The ten texts found most similar to the \textit{Questio}.}
 \label{tab:mostsimilar}
\end{table}

% \begin{table}[htp]
%  \begin{center}
%  \bgroup \renewcommand{\arraystretch}{1.3}
%  \begin{tabular}{|llc|}
%  \hline
%  \multicolumn{1}{|c}{Author} & \multicolumn{1}{c}{Document} & \multicolumn{1}{c|}{Similarity} \\
%  \hline
%  Guido Faba & Epistle 98 & 0.7024 \\
%  Guido Faba & Epistle 59 & 0.7035 \\
%  Guido Faba & Epistle 46 & 0.7116 \\
%  Guido Faba & Epistle 80 & 0.7158 \\
%  Guido Faba & Epistle 13 & 0.7266 \\
%  Guido Faba & Epistle 53 & 0.7280 \\
%  Guido Faba & Epistle 82 & 0.7288 \\
%  Guido Faba & Epistle 64 & 0.7318 \\
%  Guido Faba & Epistle 04 & 0.7450 \\
%  Guido Faba & Epistle 38 & 0.7476 \\
%  \hline
%  \end{tabular}
%  \egroup
%  \end{center}
%  \caption{The ten texts found least similar to the \textit{Questio}.}
%  \label{tab:leastsimilar}
% \end{table}%

% --------------------------------------------------------

\end{document}